\def\doctype{1}
\documentclass[sigconf]{aamas} 
\usepackage{microtype}
\usepackage[utf8]{inputenc}
\usepackage{hyperref}
\usepackage{booktabs, graphicx}
\usepackage{xcolor}
\usepackage{setspace}
\usepackage{amsmath}
\usepackage{amsfonts}
\usepackage{dsfont}
\usepackage{booktabs}
\usepackage{geometry}
\usepackage[normalem]{ulem}
\usepackage{balance} % for balancing columns on the final page
\usepackage{float}
\usepackage{subfig}
\usepackage{tikz}
\usetikzlibrary{arrows.meta,calc,decorations.markings,math,arrows.meta,patterns}
\graphicspath{{figures/}}
\usepackage[algo2e,linesnumbered,ruled,lined,resetcount]{algorithm2e}
\usepackage{longtable}
\usepackage{pdflscape}
\usepackage{xspace}

\if\doctype1
\setcopyright{rightsretained}
\acmConference[]{Copyright held by the authors}{M. Hasanbeig, D. Kroening}{H. Wijk}{M. Giacobbe}
\copyrightyear{2021}
\acmYear{2021}
\acmDOI{}
\acmPrice{}
\acmISBN{}
\else
\setcopyright{ifaamas}
\acmConference[AAMAS '21]{Proc.\@ of the 20th International Conference on Autonomous Agents and Multiagent Systems (AAMAS 2021)}{May 3--7, 2021}{London, UK}{U.~Endriss, A.~Now\'{e}, F.~Dignum, A.~Lomuscio (eds.)}
\copyrightyear{2021}
\acmYear{2021}
\acmDOI{}
\acmPrice{}
\acmISBN{}
\fi

\acmSubmissionID{616}

\title{Shielding Atari Games with Bounded Prescience}

\author{Mirco Giacobbe}
\affiliation{
  \department{Department of Computer Science}
  \institution{University of Oxford}}
\email{mirco.giacobbe@cs.ox.ac.uk}

\author{Mohammadhosein Hasanbeig}
\affiliation{
  \department{Department of Computer Science}
  \institution{University of Oxford}}
\email{hosein.hasanbeig@cs.ox.ac.uk}

\author{Daniel Kroening}
\affiliation{
  \institution{Amazon, Inc.}}
\authornote{The work reported in this paper was done prior to
joining Amazon.}
\email{daniel.kroening@magd.ox.ac.uk}

\author{Hjalmar Wijk}
\affiliation{
  \department{Department of Computer Science}
  \institution{University of Oxford}}
\email{hannes.hjalmar.wijk@cs.ox.ac.uk}

\begin{abstract}
Deep reinforcement learning (DRL) is applied in
safety-critical domains such as robotics and autonomous driving.
It achieves super-human abilities in many tasks, however whether
DRL agents can be shown to act safely is an open problem.
Atari games are a simple yet challenging exemplar for
evaluating the safety of DRL agents and feature a diverse portfolio of
game mechanics. The safety of neural agents has been studied before
using methods that either require a model of the system dynamics
or an abstraction;
unfortunately, these are unsuitable to Atari games 
because their low-level dynamics are complex and hidden inside their emulator.
We present the first exact method for analysing and
ensuring the safety of DRL agents for Atari games. Our method
only requires access to the emulator.
First, we give a set of properties that
characterise ``safe behaviour'' for several games.
Second, we develop a method for exploring all traces induced by an agent
and a game and consider a variety of sources of game non-determinism.
We observe that the best available DRL agents reliably satisfy only very
few properties; several critical properties are violated by all agents.
Finally, we propose a countermeasure that combines a bounded
explicit-state exploration with shielding.
We demonstrate that our method improves the safety of all agents
over multiple properties. 

\end{abstract}

\if\doctype1
\else
\keywords{Safe AI, Deep Reinforcement Learning, Atari Games}
\fi

\newcommand{\BibTeX}{\rm B\kern-.05em{\sc i\kern-.025em b}\kern-.08em\TeX}

\newcommand{\shield}{BPS\xspace}

\begin{document}

%%% The following commands remove the headers in your paper. For final 
%%% papers, these will be inserted during the pagination process.

\pagestyle{fancy}
\fancyhead{}

%%% The next command prints the information defined in the preamble.

\maketitle 

%%%%%%%%%%%%%%%%%%%%%%%%%%%%%%%%%%%%%%%%%%%%%%%%%%%%%%%%%%%%%%%%%%%%%%%%

\section{Introduction}

% deep rl and safety analysis

Deep reinforcement learning (DRL) combines neural network architectures 
with reinforcement learning (RL) algorithms and,
capitalising on recent advances in both technologies,
has been successfully employed in many areas of 
artificial intelligence, from playing games against humans to
controlling robots in the physical world~\cite{alphazero,alphastar,gu2016deep}.
A setup of this kind consists of an agent, a neural network, that
automatically learns
to control the behaviour of the environment by maximizing rewards received
as consequence of its actions. DRL has demonstrated super-human
capabilities in numerous applications, notably,
the game of Go~\cite{alphazero}.
DRL is now used in safety-critical domains such as 
autonomous driving~\cite{kiran2020deep}.
While DRL agents perform well most of the time,
the question of whether unsafe behavior may occur in
corner cases is an open problem.
Safety analysis answers the question of whether the environment can possibly
steer the system into an undesirable state or, dually,
whether the agent is guaranteed to remain within a set of safe states
(an invariant) in which nothing bad happens~\cite{SafeRLSurveyL,luckcuck2018formal,cautiousRL}.
We discuss the safety of popular DRL methods for one of the most challenging 
benchmark environments: the Atari 2600 console games.  

% atari games and non-determinism
Games for the classic Atari 2600 console are environments
with low-resolution graphics and small memory footprints,
which are simple when compared with contemporary games, yet offer
a broad variety of scenarios including many that are difficult for modern
AI methods~\cite{corr/MnihKSGAWR13,brockman2016openai,machado2017revisiting,AtariTrainProtocol}.
Macroscopically, diversity in the game mechanics challenges the
generality of the machine learning method; microscopically,
diversity in the outcome for multiple identical
plays, i.e., the \emph{non-determinism} in the game,
challenges the robustness of the trained agent.
Many Atari games exploit variations in the response
time of the human player for differentiating runs and, in some cases,
for initializing the seeds of random number generators.
The Arcade Learning Environment (ALE),
i.e., the framework upon which the OpenAI gym for Atari is built,
introduces non-determinism by randomly injecting no-ops, skipping frames,
or repeating agent actions~\cite{DetAtari,machado2017revisiting}. On one
hand, this prevents overfitting the agent but, on the other hand,
implies that there is no guarantee that an agent works all of the time---the
scores that we use to rank training methods are averages. Agents are
trained for strong average-case performance.

\begin{figure}[!t]
	\centering
	\input{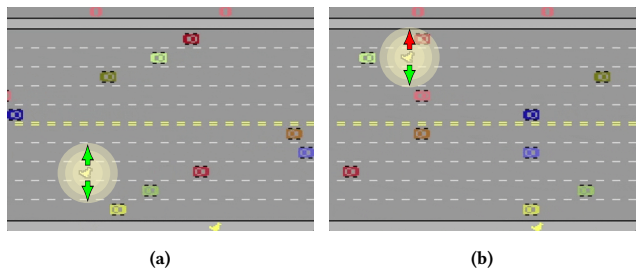}
	\caption{The effect of the bounded-prescience shield on the game \texttt{Freeway}.
		In (a), both actions `\texttt{up}' and `\texttt{down}' are safe
		and allowed; in (b), the action `\texttt{up}' is unsafe and blocked
		by the shield.}
	\label{fig:freeway}
\end{figure} 

The application of DRL in safety-critical applications, by contrast,
requires worst-case guarantees, and we expect a safe agent to
maintain \emph{safety invariants}. To evaluate whether or not
state-of-the-art DRL delivers safe agents we specify a collection
of properties that intuitively characterize safe behaviour for a
variety games, ranging from generic properties such
as ``don't lose lives'' to game-specific ones such as avoiding
particular obstacles. Figure~\ref{fig:freeway} illustrates
the property ``duck avoids cars'' in the game Freeway.
In the scenario in Fig.~\ref{fig:freeway}a this property is maintained
regardless of the action chosen by the agent whereas the
scenario given in Fig.~\ref{fig:freeway}b offers the possibility
of violating it.  We conjecture that satisfying our properties is
beneficial for achieving a high score, and therefore study whether
neural agents trained using best-of-class DRL methods learn to
satisfy these invariants. Finally, we discuss a countermeasure
for those that violate them.

The safety of DRL has been studied from the
perspective of verification,
which determines whether an trained agent is safe as-is~\cite{cav/HuangKWW17}, 
and that of synthesis,
which alters the learning or the inference processes
in order to obtain a safe-by-construction agent~\cite{SafeRLSurveyL}. 
Verification methods for neural agents have borrowed
from constraint satisfaction or abstract
interpretation~\cite{katz2017reluplex,nips/BunelTTKM18,sp/GehrMDTCV18}. 
%
%% The first solve the problem of finding an adversarial input
%% for which the agent violate a constraint for the safe states \cite{};
%% the second construct a symbolic overapproximation of the
%% reachable states of the system and decide whether these are safe \cite{}.  
%
Both approaches are symbolic and, for this reason, require
a symbolic representation not only for the neural agent
but also for the environment.
They have been used for reasoning about
neural networks in isolation, e.g., image classifiers~\cite{cav/HuangKWW17,cav/HuangKWW17}, or
for environments whose dynamics are determined by symbolic expressions,
e.g., differential equations~\cite{cav/HuangKWW17,cav/TranYLMNXBJ20}; unfortunately, they are
unsuitable to Atari games because their mechanics are
hidden inside their emulator, Stella, i.e., the core of ALE.
For this reason, we adopt an \emph{explicit-state} verification strategy and
then, building upon it, we construct safe agents.

We introduce a novel verification method for neural agents
for Atari games. Our method explores all reachable states
explicitly by executing, through ALE, 
the games and agents and labels each state for whether
it satisfies our properties.
More precisely,
we enumerate all traces induced by the non-deterministic
initialisation of the game
and label states using their lives count, rewards,
and the screen frames generated,
which allows us to specify 43 non-trivial properties
for 31 games.
We compare agents trained using different technologies, i.e.,
A3C~\cite{mnih2016asynchronous}, Ape-X~\cite{horgan2018distributed}, DQN~\cite{NatureDQN}, IQN~\cite{dabney2018implicit}, and Rainbow~\cite{Rainbow},
and observe that all of them violate 24 of our
properties, whereas only 4 properties are satisfied by all.
Surprisingly, properties that are intuitively difficult for humans,
e.g., not dying, are satisfied by some agents,
whereas many that we judge as simple, e.g.,
keeping a gun from overheating in game Assault, are violated by all agents. 
To improve the overall safety of neural agents wrt.~our properties,
we employ our explicit-state labelling and exploration technique
to \emph{shielding} neural agents.

Ensuring safety amounts to constraining the traces of the
system within those that are admissible by the safety property.
Methods that act on the training phase modify the
optimization criterion or the exploration process in order to
obtain neural agents that naturally act safely~\cite{SafeRLSurveyL}. 
Methods of this kind typically require known facts about the environment
for providing guarantees and have not been applied to Atari games,
or exploit external knowledge (e.g., teacher advice)~\cite{ScaleHuman}.
On the other side of the spectrum,
shielding enables the option of fixing unsafe
agents at inference phase only, introducing a third
actor---the shield---that takes over control when necessary and with minimal
interference~\cite{ShieldBeforeAfter,ShieldBeforeAfter_2,ShieldBeforeAfter_3}.
A shield is constructed from a safety property in temporal logic
and a model of the environment or an abstraction.
Leveraging the fact that the safety property is usually easy
to satisfy in contrast to the main objective,
shielding is efficient with respect to training for safety.
However, complete models for Atari games are not available and
abstractions are hard to construct automatically; 
for this reason, we adapt shielding to our exploration method.

We study the effect of shielding DRL agents from actions
that lead to unsafe outcomes within some bounded time in the future.
For this purpose, we augment agents with shields that, during execution,
restrict their actions to those that are necessarily safe within
the prescience bound.
Before taking an action, our \emph{bounded-prescience shield} (BPS)
enumerates all traces from the current state for a bounded number of steps
and labels each of them as safe or unsafe using our verification technique;
then, it invokes the agent and chooses the next action whose traces are all
labelled as safe and whose agent score is the highest. 
As a result, we fixed all violated properties that we deemed as simple
using BPSs with shallow prescience bounds of 3 steps.
Notably, we also fixed the properties that we consider non-trivial and
that were satisfied by most non-deterministic executions under the original agent.
Overall, BPS demonstrated its effectiveness for those properties that
are simple yet always violated by the original agents, or those that
are difficult yet were almost satisfied.

Summarising, our contribution is threefold.
First, we enrich the Atari games with the first comprehensive library of
specifications for studying RL safety.
Second, we introduce a novel technique for evaluating the safety of agents
based on explicit-state exploration and discover that current DRL algorithms
consistently violate most of our safety properties.
Third, we propose a method that, exploiting bounded foresight of the future,
has mitigated the violation of a set of simple yet critical properties,
without interfering with the main objective of the original agents. 
To the best of our knowledge, our method has produced
the safest DRL agents for Atari games currently available. 

\section{Safety for Atari Games}\label{sec:atari}

In this section we discuss ALE~\cite{machado2017revisiting}, which is a
tool for running Atari 2600 games based on the Stella
emulator.  While any of the hundreds of available Atari games can be loaded
into the emulator, ALE provides built-in support for 60 games and
those are generally the ones studied.  This set of games contains a wide
variety of different tasks and dynamics.

\subsection{Markov Decision Processes}

We focus on the standard formalisation of sequential decision-making
problems, i.e., Markov decision processes (MDP), which assumes that the
actions available, the rewards gained and the transition probabilities only
depend on the current state of the environment and not the execution
history.  Formally, an MDP is given as a tuple $\mathcal{M}=(S,s_0,A,P,R)$,
where $S$ is the set of states of the environment, $s_0$ is the initial
state, and $A$ is the set of actions.  The dynamics of the environment are
described by $P: S\times A \times S\to [0,1]$, where $P(s,a,s')$ is the
probability of transitioning to $s'$ given the agent chooses action~$a$ in
state~$s$.  The obtained reward when action~$a$ is taken in a given
state~$s$ is a random variable
$R(s,a)\sim\rho(\cdot|s,a)\in\mathcal{P}(\mathbb{R})$, where
$\mathcal{P}(\mathbb{R})$ is the set of probability distributions on subsets
of $\mathbb{R}$, and $\rho$ is the reward distribution.  A possible
realisation of $R$ is denoted by $r$~\cite{puterman2014markov}.

Partially observable Markov decision processes (POMDPs) are general cases of
MDPs, and Atari games can perhaps be most naturally modelled as a POMDP. 
When defining a POMDP, the MDP tuple $\mathcal{M}=(S,s_0,A,P,R)$ is extended
with a set of \textit{observations} $\Omega$ and a conditional observation
probability function $O:S \times A\times \Omega\to \mathbb{R}$.  When
picking an action~$a$ in state~$s$, the agent cannot observe the
subsequent state $s'$ but instead receives an observation $o\in\Omega$ with
probability~$O(o|s',a)$.  Unlike when using MDPs one cannot assume that an
optimal policy for a POMDP will depend only on the last observation---in
fact, effective use of memory is often crucial.  Further, we assume that the
state-space $S$ includes a ``terminating state''.

In an Atari game the full state $s\in S$ is given by a valuation of the
$128$-byte RAM, along with a set of registers and timers.  There is no
additional screen buffer, which means an observation $o\in \Omega$ is given
by a $210\times160$ display frame, which is computed deterministically from
the state~$s$.  ALE executes the Atari games using the Stella
emulator, and treats this emulator almost entirely as a black-box.  The only
manipulation ALE carries out during the run of a game is sending the control
input selected by the agent to the emulator, reading the screen and reading
two fixed memory addresses where the score (used as the reward signal) and
the current number of lives are stored.  There are a total of $18$ discrete
actions possible in any state, including ``no operation''.  The Atari games
are all deterministic.  This essentially means that the above POMDP is
easily convertible into an MDP where at each time step $t$, the MDP state
$s_t$ is a finite sequence of observations and actions,
i.e.~$s_t=o_0,a_0,o_1,a_1,\ldots,o_t,a_t$.  This formalization gives rise to
a large but finite MDP.

\subsection{Safety Properties}

Traditionally the reward signal exposed by ALE is used as the only measure
of success.  In order to study to which degree the behavior of trained
agents is safe, we hand-engineer a suite of 43 safety properties across 30
games, which identify unsafe states of the MDP.  The choice of the
properties is highly subjective.  The authors believe that all properties
should be satisfied at all times by a highly reliable and robust agent. 

We observe that some of the properties are easy to satisfy whereas others
require near perfect gameplay.  Consider the Atari game Bowling.  The
property \texttt{\textbf{Bowling:no-strike}} identifies any state in which
the player fails to score a strike as unsafe.  We also include ``not losing
lives'' as a property in all games where it applies, and this property can also
be highly challenging in many games.  To better interpret the results we
identify two distinct sets of properties that we consider ``easy'' and
``hard''.

\begin{figure}
\centering
\includegraphics[scale=0.7]{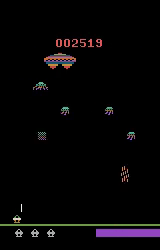}
\caption{A frame from the Atari game Assault.  The heat bar (purple,
lower right) is almost full, and another shot fired in this state would lead
to overheating, violating \texttt{\textbf{Assault:overheat}}.}
\label{fig:overheat}
\end{figure}

\begin{figure}
	\centering
	\subfloat{
		\includegraphics[width=.35\linewidth]{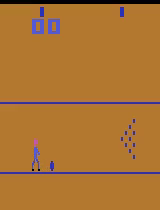}
	}
	\subfloat{
		\includegraphics[width=.35\linewidth]{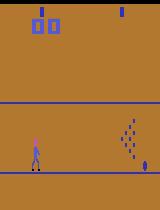}
	}
	\caption{A violation of \texttt{\textbf{Bowling:no-hit}}, where the player misses all pins.}
	\label{fig:bowling}
\end{figure}

\subsubsection{Shallow Properties}

We say that a property is \textit{shallow} if violations of the property are
always caused by recent actions (within 10 frames or fewer).  More
precisely, for any unsafe state encountered during a trace, there should be
a previous state at most 10 frames earlier in the trace from which a safe
strategy exists.  In \texttt{\textbf{Assault}}, the player loses a life from
overheating if they overuse their weapon in a short timespan.  The property
\texttt{\textbf{Assault:overheat}} marks states where such an overheating
happens as unsafe.  This is an example of a shallow property, since ``not
firing'' is always a safe strategy starting from the frame just before
overheating (such a frame is given in Figure~\ref{fig:overheat}).  Another
example is the game \texttt{\textbf{DoubleDunk}}, where the player
is penalised for stepping outside of the field.  This violates the property
\texttt{\textbf{DoubleDunk:out-of-bounds}}, and whenever a violation occurs,
simply moving in the opposite direction a few frames back would have avoided
the violation.

Ensuring the safety of shallow properties does not require long-term
planning. It is reasonable to expect that most agents will satisfy all
shallow properties.

\subsubsection{Minimal Properties}

We say that a property is \textit{minimal} if satisfying it is a necessary
requirement for scoring at least 10\% of the human reference level. 
Violating any of these properties would indicate a complete inability to play
the game, and yield a near-zero or negative score.  An example is
\texttt{\textbf{Bowling:no-hit}}, which marks states where all pins are
missed as unsafe.  We observe that while most minimal properties are usually
``easy'', they are not necessarily shallow as can be seen in
Figure~\ref{fig:bowling}.  By the time the miss occurs the throw that caused
it happened hundreds of frames in the past.

\section{Deep RL Algorithms}

It has been shown~\cite{puterman2014markov,puterman2} that in any MDP
$\mathcal{M}$ with a bounded reward function and a finite action space, if
there exists an optimal policy, then that policy is stationary and
deterministic, i.e.  $\pi:S\to A$.  A~deterministic policy generated by a
DRL algorithm is a mapping from the state space to action space that
formalises the behaviour of the agent whose optimisation objective is
$$\mathbb{E}[\sum_{t=0}^{\infty}\gamma^t R(s_t,~a_t)],$$
where $0<\gamma\leq1$ is the discount factor, and the goal is to find a
policy, call it $\pi^*$, that maximises the above expectation.

This paper focuses on model-free RL due to its success when dealing with
unknown MDPs with complex dynamics including Atari games~\cite{NatureDQN},
where full models are difficult to construct.  A~downside of model-free RL
however is that without a model of the environment formal guarantees for
safety and correctness are often lacking, motivating the work on safe
model-free RL~\cite{cautiousRL}.  A classic example of model-free RL is
Q-learning (QL)~\cite{watkins1992q}, which does not require any access to
the transition probabilities of the MDP and instead updates an action-value
function $Q:S\times A\to \mathbb{R}$ when examining the exploration traces. 
While vanilla model-free RL, e.g.~QL, avoids the exponential cost of fully
modelling the transition probabilities of the MDP, it may not scale well to
environments with very large or even infinite sets of states.  This is
primarily due to the fact that QL updates the value of each individual
state-action pair separately, without generalising over similar state-action
pairs.  To alleviate this problem one can employ compressed approximations
of the $Q$-function.  Although many such function approximators have been
proposed~\cite{cmac,ormo,ernst,babuska,baird1995residual} for efficient
learning and effective generalisation, this paper focuses on a particular
representation, which has seen much success in recent years: neural
networks~\cite{apprx}.

Neural networks with appropriate activation functions are known to be
universal function approximators when given enough hidden
units~\cite{csaji2001approximation, apprx}.  Thus, the use of DNNs with many
hidden layers requires only few assumptions on the structure of the problem
and consequently introduces significant flexibility into the policy.  More
importantly, it has been shown empirically that despite being severely
overparametrised, neural networks seem to generalise well if trained
appropriately.  This, along with efficient algorithms for fitting networks
to data through backpropagation, has made the use of DNNs widespread.  In RL
this has led to the paradigm of DRL~\cite{arulkumaran2017deep}.

The performance of DRL when applied to playing Atari 2600 games using
raw image input led to a surge of interest in DRL~\cite{NatureDQN}.  The
$Q$-function in~\cite{NatureDQN} is parameterised $Q(s,a|\theta)$, where
$\theta$ is a parameter vector, and stochastic gradient descent is used to
minimise the difference between $Q$ and the target estimate and by
minimizing the following loss function:
\begin{equation}\label{eq:loss_function}
L(\theta)= \mathbb{E}_{(s,a,r,s') \sim \mathcal{U}}[(r+\gamma\max\limits_{a'\in A}Q(s',a'|\theta)-Q(s,a|\theta)^2].
\end{equation}
Actions and states are sampled by letting the agent explore the environment.
The experience data points $(s,a,r,s')$ are stored in a replay
buffer~$\mathcal{U}$, as in vanilla QL.  Generally this pool of
experiences is capped at a certain size at which point old ones cycle out. 
This means the training objective evolves only gradually and additionally
ensures that individual experiences are more independent rather than always
being consecutive, reducing the variance in training.  This however exposes
an important difference to regular supervised learning, where the updates
that are made to the $Q$-function change the distribution of data, and thus
also the training objective.  This means training the neural network to
correctly estimate $Q$-values given the current policy has to be interleaved
with gathering new data by letting the agent interact with the environment.

Convergence of neural-network-based methods is in general much less certain
than it is for QL that uses a look-up table.  There are many other methods
for DRL that are not closely based on QL.  In particular, there are policy
gradient approaches that do not attempt to estimate the value of states but
rather directly fit policy parameters to maximise rewards.  The common
aspect of these methods is that they scale well to large and complex
problems, but in turn are often opaque and lack theoretical
convergence guarantees.

We gathered $29$ state-of-the-art DRL algorithms for Atari games to find the
policies that achieve the best performance across all games:
\begin{itemize}

\item We include all algorithms whose implementations are available for
ALE~\cite{machado2017revisiting} or OpenAI Gym~\cite{brockman2016openai}.

\item We consider every algorithm that achieved a top five score on any of
the OpenAI Gym
leader-boards\footnote{\href{https://github.com/openai/gym/wiki/Leaderboard}{https://github.com/openai/gym/wiki/Leaderboard}}.

\item We additionally include algorithms that are prominently benchmarked
against by other works.

\end{itemize}

\begin{figure}[!t]
	\centering
	\includegraphics[width=0.9\columnwidth]{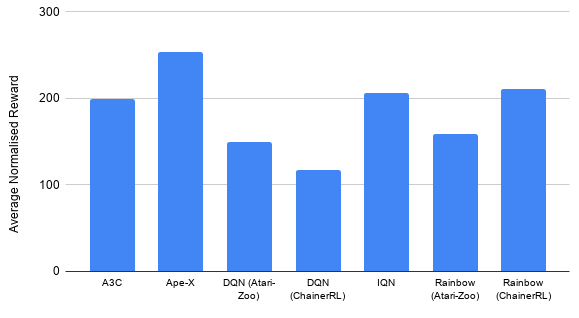}
	\caption{Average normalised reward obtained over all games by each algorithm.}
	\label{fig:avg_reward_normalised}
\end{figure}

We ranked the algorithms by the number of games in which they placed among
the top 5 of the 29 total gathered Atari algorithms.  To avoid mistakes in
training from affecting our assessment we then restricted the search to
algorithms for which benchmarked pretrained agents were available.  This
ultimately gave us the list of top performing available algorithms:
 Ape-X~\cite{horgan2018distributed},
 A3C~\cite{mnih2016asynchronous},
 IQN~\cite{dabney2018implicit}, and
Rainbow~\cite{Rainbow}. We also included the traditional DQN~\cite{NatureDQN}
for comparison.  Figure \ref{fig:avg_reward_normalised} shows the
average normalised rewards $R_n$ over all games for each algorithm in our testing where
$$R_n=100\times\frac{R-R_r}{R_h-R_r},$$
where $R_r$ is the average reward of a random agent, and $R_h$ a recorded
average of human play~\cite{NatureDQN}. We use a shorter episode length
(5\,min.) than that used in~\cite{NatureDQN} to be able to run more traces,
so the result is not directly comparable (e.g.  100\% is above human level,
since it is accomplished in less time).  We still apply the normalisation
since the purpose is not to compare with humans or other studies but to
normalise the relative impact of each game within our study.

The main question is whether these top performing algorithms are able to
satisfy safety properties that are obviously desirable to a human player. 

\section{Safety Analysis via Explicit-state Exploration}

We are interested in proving invariant safety properties, i.e., we want to
show that the synthesised policy never enters a state that is labelled
`unsafe'.  The definition of a property thus boils down to defining a set of
unsafe states, or equivalently its complement, a set of safe states.  Let
$\varphi$ be a safety property whose labelling function is denoted by
$L_\varphi:S\to \{\text{safe},\text{unsafe}\}$.  With $s_0$ being the
starting state of the Atari game, we define $s_{i+1}=P(s_i,\text{no-op})$ as
the sequence of states achieved by repeatedly performing the `no-op' action. 
We then have $I=\{s_i|i<\nu\}$ as the set of initial states from which an
action selection policy $\pi$ is followed.  We assume the state-space $S$
contains a terminating state $\bot$, which is always labelled safe.

To verify $\varphi$, i.e.  that the system will never enter an unsafe state,
we simply need to check whether any reachable state has label `unsafe'. 
Given that all the games and the agents are deterministic except for having
multiple starting states we can simply follow the deterministic path
starting at each initial state $s\in I$ until we reach the terminating
state, from which no other state is reachable.  We record if the labelling
function associated with $\varphi$ reports `unsafe' for any state on the
path.

\subsection{Non-determinism in Atari Games}

One of the more difficult aspects of training and evaluating models on Atari
is appropriately handling non-determinism.  The dynamics within each game
are entirely deterministic, other than the initialisation behaviour, which
depends on an adjustable seed.  The Stella emulator very closely emulates
the original hardware and also performs random RAM initialisation using
a seed that is derived from the system clock.  In order to bring
back some of the intended randomness of the original games, and also to
create a more interesting training environment that requires some level of
generalisation, ALE introduces additional forms of stochastic behaviour. 
Since the environments in ALE are treated in a black-box manner this is done
purely through modifying the actions selected by the policy.  Some of the
most common ways of introducing stochasticity include:
\begin{itemize}
	
	\item \textbf{No-ops}, where ALE sends a random number between $0$
	and $30$ of no-operation actions at the start of the game, both letting the
	environment evolve into a random starting state and randomly seeding the
	game.
	
	\item \textbf{Sticky actions}, where a random chance (often 25\%) of
	repeating the previous action is introduced every frame.  This in some sense
	mimics a human's imperfect frame timing, since a human player is not able to
	trigger an action in sync with a particular frame reliably.
	
	\item \textbf{Frame skips}, where each action is repeated a random
	number of times, e.g.  in OpenAI Gym \cite{brockman2016openai} the default
	is between $3$--$5$ times.  This is very similar to sticky actions, but with
	a different probability distribution over number of times the action is
	repeated.  Importantly, frame-skips have finite support, e.g.  with Gym-style
	frameskips there is an equal $\frac{1}{3}$ probability of $3$, $4$ or $5$
	repeats and no chance of any other number, whereas sticky actions can in
	theory lead to repeating the action an arbitrary number of times before
	giving back control to the policy.
	
	\item \textbf{Human starts}, which is a more elaborate version of
	the no-ops start where ALE sends a memorised series of commands based on a
	human trace before handing over control to the policy.
	
\end{itemize}

Each method has distinct advantages and disadvantages. No-ops and also human starts randomise over
a certain fixed number of starting conditions
\cite{DetAtari,machado2017revisiting} while ALE and OpenAI Gym adopted sticky actions and frame skips respectively.

\subsection{Labelling Functions}

The labelling function $L_\varphi$ can be defined as a mapping directly from the
underlying machine state (RAM).  However, as stated before, correctly
interpreting the RAM to define even simple properties proved to be
difficult.  We thus use the history of actions, video frames, the life
counter and rewards as the state passed to the labelling functions instead. 
We categorise the labelling functions into three classes:
\begin{itemize}

	\item \textbf{Life-count Labelling:} A common safety property that
is used across many games is simply avoiding losing a life or
$\varphi=\texttt{\textbf{dying}}$. For
games where there is a life counter, the Atari 2600 emulator returns the
number of lives left in the game~\cite{machado2017revisiting}.  A labelling
function for these properties is easy to define.  Namely, the
labelling function labels a state unsafe if the life counter reported by ALE
is reduced compared to the previous state.

	\item \textbf{Reward-based Labelling:} Another set of labelling
functions are those that are directly induced from the game score.  For
instance, a safety property in \texttt{\textbf{Boxing}} is `not to get
knocked out' ($\varphi=\texttt{\textbf{no-enemy-ko}}$): the agent gets
knocked out if the opponent scores $100$ hits on the agent.  Since there is
no other way of losing score, a function that only accumulates the total
negative reward labels a state as unsafe once $-100$ is reached.  Many other
properties can be derived from the reward through various schemes similar to
\texttt{\textbf{Boxing}}.

	\item \textbf{Pixel Image Labelling:} Some safety properties however do not
correspond clearly to any specific reward or life-loss signal.  For
instance, $\varphi=\texttt{\textbf{overheat}}$ in \texttt{\textbf{Assault}}
results in the exact same punishment
as dying, however avoiding overheating represents a distinct and easier
behaviour than avoiding death in general.  To label such properties we
process raw RGB frames and examine pixels of specific colours in specific
places, or track the position of objects on the screen.  The simplistic
graphics of the Atari 2600 makes the image processing and labelling
real-time.  However, this type of labelling functions requires by far the
most work and is also most prone to mistakes.

\end{itemize}

\subsection{Safety Analysis Results}

We initialise each game with $\nu=30$ rounds of no-ops of different length;
each round produces a different initial state.
For each of these initial states,
we run the game together with the agent and recorded
whether an unsafe state was eventually reached and, if so,
how many steps it takes. Additionally, we record the total reward
achieved by the policy over the trace.
As a result, for each game we obtain whether it satisfies the property,
that is all traces satisfy it, or additionally measured the degree of safety,
determined by the ratio of satisfying traces over all traces.

Overall we run 301 analyses, 43 for each of the 7 algorithms,
out of which 72 show an agent satisfying a safety property.
%% \begin{figure}[!t]
%%   \centering
%%   \includegraphics[width=0.7\columnwidth]{safetyanalysis/sat_by_algo}
%%   \caption{Satisfied properties by RL algorithm.}
%%   \label{fig:sat_by_algo}
%% \end{figure}
\begin{figure}[!t]
  \centering
  \includegraphics[width=0.9\columnwidth]{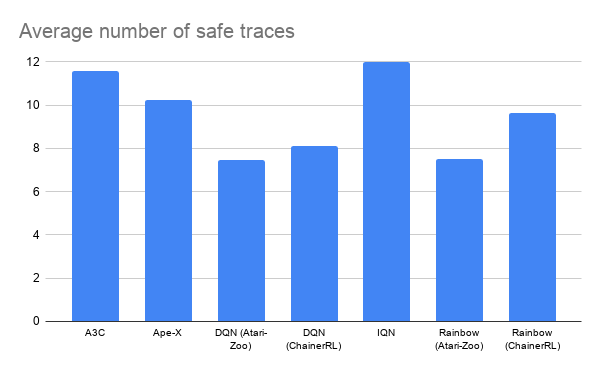}
  \caption{Number of safe traces by RL algorithm.}
  \label{fig:avg_safety_by_algo} 
\end{figure}
Figure~\ref{fig:avg_safety_by_algo} gives the performance
of the agent trained by each
algorithm with respect to the properties it could satisfy.
Notably, IQN yields the largest number of safe traces,
followed by A3C in second and Ape-X in third place.
%% Asynchronous Actor-Critic\cite{mnih2016asynchronous} ,
%% followed by Ape-X\cite{horgan2018distributed} and IQN \cite{dabney2018implicit}.
%% However, their performance changes when each trace is evaluated intependently
%% as we illustrate in Fig. \ref{fig:avg_safety_by_algo}. IQN obtained the
%% highest average of safe traces per property,   

\begin{figure}[!t]
	\centering
	\includegraphics[width=0.9\columnwidth]{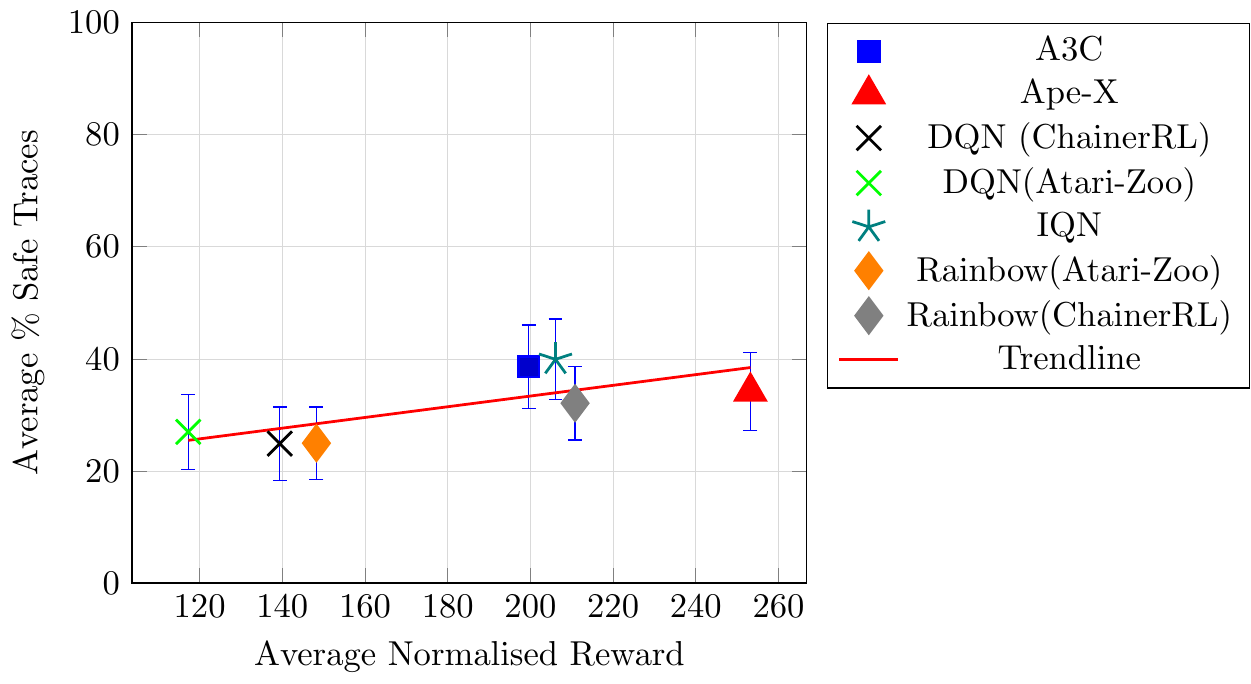}
	\caption{Correlation of safe traces and reward (r = 0.69).}
	\label{fig:correlation}
\end{figure} 

The degree of safety correlates well with the reward obtained, as we show in
Fig.~\ref{fig:correlation}.  Despite this good correlation between the
average reward and safety, all of the algorithms violate at least some
safety properties across all the games.  In absolute terms, no algorithm
achieves a 50\% safety score.  This essentially means that maximising the
reward is not equivalent to acting safe, and it is clear that the algorithms
considered do not reliably learn safe behaviour.  By inspecting individual
traces, it often appears that the agents are capable of satisfying the
safety properties, meaning that there exist examples among the traces of the
agent in which the agent correctly dealt with complex situations with high
risk of violating the safety property.  But the reward structure or perhaps
just insufficient training means it lacks reliability and robustness, and
sometimes fails even in simple situations.

Furthermore, there are still noticeable differences between the trained
agents, both comparing ones trained using different learning algorithms, and
those which only differed in implementation, e.g.  DQN Atari-Zoo and
ChainerRL.  This provides further evidence that the safety of these methods
depends on various contingent, opaque and poorly understood factors.  In
what follows we examine the properties are satisfied and violated by the
agents in more detail, with a view towards improving their safety.

\begin{figure}[!t]
  \centering
  \includegraphics[width=0.7\columnwidth]{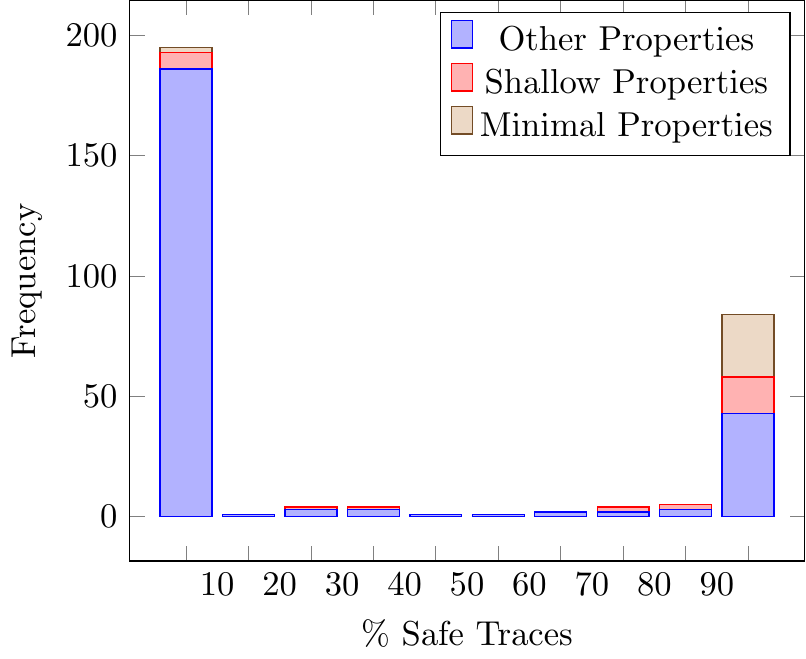}
  \caption{Distribution of safety.}
  \label{fig:distr_prop}
\end{figure}

To determine how agents behave with respect to our properties, we study the
distribution of traces satisfied by each of them. 
Figure~\ref{fig:distr_prop} illustrates how many analyses ended with all 30
runs satisfying the property, no runs satisfying the property, and
everything in between.  Notably, most outcomes are distributed in the
extreme cases, which indicates that statistically agents either satisfy a
property or they don't.  Only a small amount of cases are close to
satisfaction or close to violation.  This indicates that our safety
properties are robust and insensitive to any non-deterministic noise
emerging from the combination of game and agent.

\begin{figure}[!t]
  \centering
  \includegraphics[width=0.7\columnwidth]{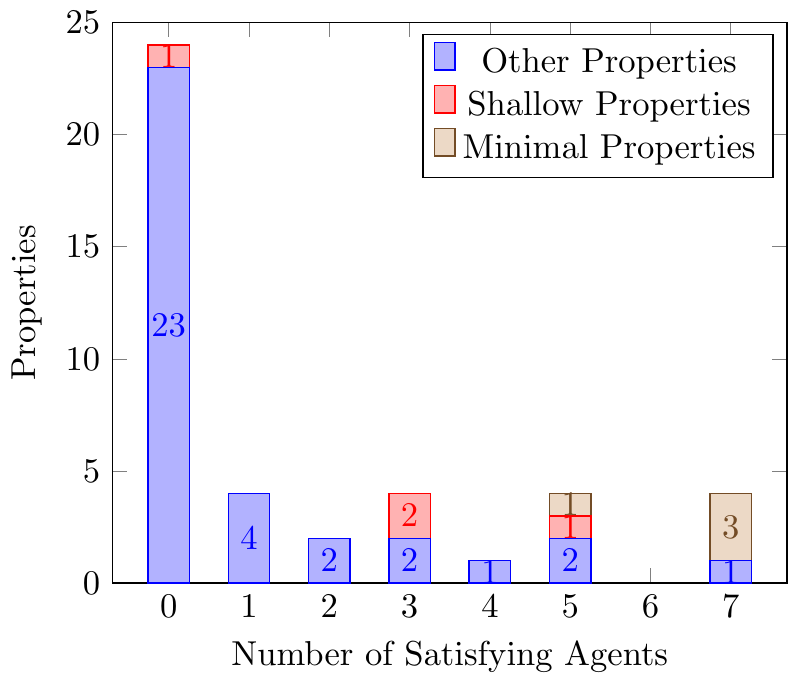}
  \caption{Distribution of agents satisfying a property.}
  \label{fig:distr_sat}
\end{figure} 

Most of our properties are consistently violated. 
Figure~\ref{fig:distr_sat} shows that 24 out of 43 of our properties are
violated by all agents.  On the other side, only 4 properties are satisfied
by all and 3 out of 4 of these properties are classified as minimal. 
Minimal properties are not easy yet essential for making progress in the
game.  Training algorithms optimize for reward and, indirectly, for progress
and therefore they satisfy minimal properties as a side effect.  This
indicates that reward functions focus on progress but are incomplete with
respect to safety.  Three shallow properties, which are properties that are
simple to satisfy, are not satisfied by all agents and one in particular is
violated by all.  These violations should be avoidable with a shallow
exploration of the future.  We investigate this hypothesis in the next
section.

\section{Bounded-Prescience Shielding}

Standard safe policy synthesis in formal methods requires full knowledge of
environment dynamics or the ability to construct an abstraction of
it~\cite{receding,raman2014model,soudjani2014faust}.  In practice, however,
the dynamics are not fully known, and abstractions are too hard to compute.

RL and DRL methods address the computational inefficiency of
safe-by-construction synthesis methods, but on the other hand cannot offer
safety guarantees~\cite{LogicGuidance,plmdp}.  This issue becomes even more
pressing when the learning algorithm entirely depends on non-linear function
approximation to handle large or continuous state-action
MDPs~\cite{hasanbeig2020deep,lcnfq,deepsynth}.  For instance, the loss
function~\eqref{eq:loss_function} in DRL and consequently the synthesised
optimal policy $\pi^*$ only accounts for the expected reward.

The concept of shielding combines the best of two worlds, that is, formal
guarantees for a controller with respect to a given property and policy
optimality despite an environment that is unknown
a~priori~\cite{ShieldBeforeAfter, ShieldBeforeAfter_2, ShieldBeforeAfter_3}. 
The general assumption is that the agent is enabled to observe the MDP and
the actions of any adversaries to the degree necessary to guarantee that the
system remains safe over an infinite horizon.

In this work, we propose a new technique we call \shield, which only
requires observability of the MDP up to a bound $H \in \mathds{N}$.  This
relaxes the requirement of full observability of MDP and adversaries and
more importantly, allows the shield to deal with MDPs with large state and
action spaces.  In particular, we will show that \shield is an effective
technique for ensuring safety in Atari games where the MDP induced by the
game is hard to model or to abstract (Section~\ref{sec:atari}).

A finite path $\rho=(s,a)$ starting from $I$ is a sequence of states and actions
$$
\rho= s_0 \xrightarrow{a_0} s_1 \xrightarrow{a_1} ... \xrightarrow{a_{n-1}} s_n
$$
such that every transition $s_i \xrightarrow{a_i} s_{i+1}$ is allowed in MDP
$\mathcal{M}$, i.e.  $P(s_i,a_i,s_{i+1})>0$ and $s_n$ is a terminal state. 
A \textit{bounded} path of length $\mathcal{L}$ is a path with no more than
$\mathcal{L}$ states, where \textit{either} the final state $s_n$ is
terminal or the number of states is exactly $\mathcal{L}$.  We denote the
set of all finite paths that start at an arbitrary state $s_p \in S$ by
$\varrho(s_p)$, and the set of all bounded finite paths of length
$\mathcal{L}$ that start at $s_p$ by $\varrho_{\mathcal{L}}(s_p)$.

Given a safety property $\varphi$, a (bounded) finite path $\rho$ is called
\emph{safe} with respect to $\varphi$, written $S(\rho,\varphi)$, if
$$
	L_\varphi(s_i)=\text{safe} ~~\forall s_i \in \rho,
$$
where $L_\varphi:S\to \{\text{safe},\text{unsafe}\}$ is the labelling
function of the safety property $\varphi$.  A policy $\pi$ is safe with
respect to a property $\varphi$ if for any state $s_0$,
\begin{align}\label{eq:safe_prop}
	&\exists \rho=(s,a)\in\varrho(s_0) (S(\rho,\varphi) \wedge \pi(s_0)=a_0) \vee \\
	\nonumber  &\forall \rho\in\varrho(s_0)(\neg S(\rho,\varphi)),
\end{align}
or in other words if the policy always picks an action that starts a safe
finite path if one exists.  Finally, a policy is bounded safe with bound~$H$
with respect to $\varphi$ if Equation~\ref{eq:safe_prop} is satisfied
when replacing ``finite paths'' by ``bounded paths of length~$H$''.

\begin{figure}[!t]
  \includegraphics[width=.9\columnwidth]{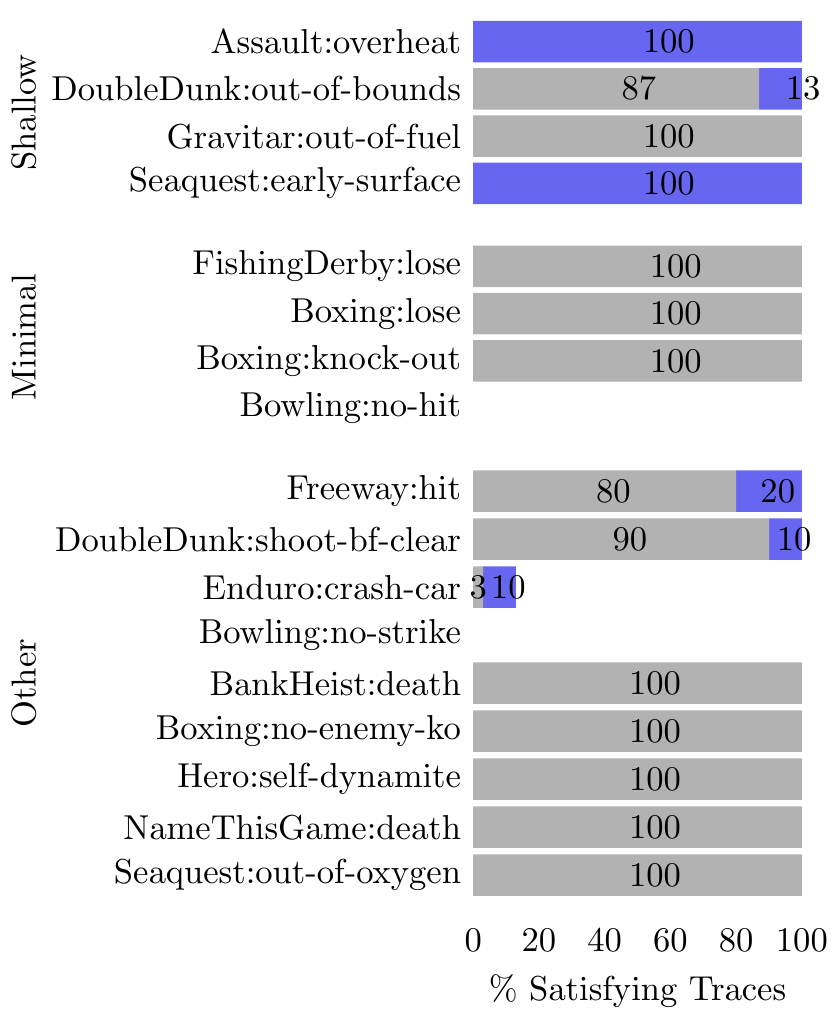}
  \caption{Additional Safety gained by applying a BPS with bound 5 to DQN. The other 26 properties were safe in 0\% of traces with and without shielding and are not shown due to space limitations.}
  \label{fig:DQNshield}
\end{figure}
\begin{figure}[!t]
  \centering
  \includegraphics[width=.9\columnwidth]{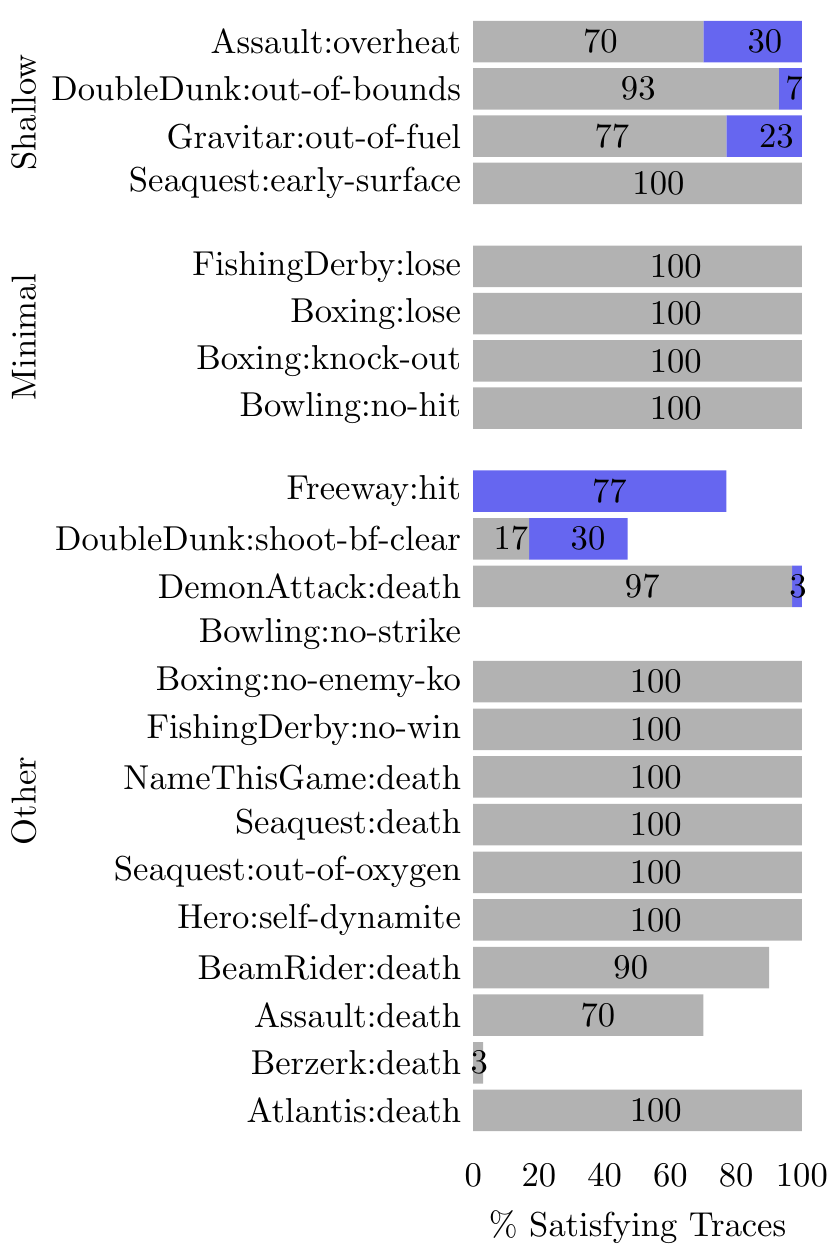}
  \caption{Additional Safety gained by applying a BPS with bound 5 to IQN. The other 21 properties were safe in 0\% of traces with and without shielding and are not shown due to space limitations.}
  \label{fig:IQNshield}
    %\label{Effect of applying a BPS with bound 5 on safety.}
\end{figure} 

Our shield modifies the policy $\pi$ of the trained agent to obtain a
policy~$\pi'$ that is guaranteed to satisfy bounded safety.  This is done by
forward-simulating $H$ steps and forbidding actions that cannot be continued
into a safe bounded path of length~$H$.  Where the policy has preferences
among available actions we pick the most preferred one that starts some
safe bounded path.  This is the case for all our DRL agents, whose final
layer expresses full preferences.  If none of the available actions start a
safe bounded path, the shield reverts to the original policy~$\pi$ (still
satisfying bounded safety, by the second operand of~\ref{eq:safe_prop}).

In the worst case this requires enumerating all bounded paths
in~$\varrho(s)$ before finding a safe one, and if $n$ actions are available
from each state there will be up to $n^H$ such bounded paths, which can make
shielding with large bounds~$H$ intractable.  In practice, unsafe states are
relatively rare and a safe path can be found quickly from most states.  In
particular, $\pi$ itself will often be bounded safe for most states, and can
be followed directly.

By guessing $\pi$ is safe and rolling back to explore other paths only if a
violation occurs, our algorithm has minimal computational overhead as long
as $\pi$ continues being safe.  By also remembering unsafe paths between
time-steps, the shield can become performant even when encountering
violations, for small bounds $H$, as evaluated in
Figure~\ref{fig:shield_compute}.

%\begin{center}\commentt{$\leftrightsquigarrow \leftrightsquigarrow \leftrightsquigarrow$ Hosein: edited up to here $\leftrightsquigarrow \leftrightsquigarrow \leftrightsquigarrow$}\end{center}

\subsubsection*{Experimental Results}

We evaluated the effectiveness of \shield on robustifying DQN and IQN
against the safety properties including \emph{Shallow} and \emph{Minimal}
properties.  Recall that shallow properties are those that the we expect to
need a prescience bound with 10 frames or fewer, and minimal properties are
those that are necessary for scoring $10\%$ of human game-play level.

Fig.~\ref{fig:DQNshield} and Fig.~\ref{fig:IQNshield} illustrate the
performance of the DQN- and IQN-trained agents before and after applying
BPS.  The prescience bound of the shield in both experiments is $H=5$.  We
initialised each \texttt{\textbf{game:property}} with $\nu=30$ rounds of
no-ops and monitored safety violations over all $30$ generated traces.  Note
that applying \shield significantly improved the performance of both
algorithms in Shallow properties, and with no further training both DQN and
IQN fully satisfied the safety properties.  This comes at a minimal
computation cost as compared to re-training DRL algorithms to achieve the
same performance.

\begin{figure}[!t]
  \centering
  \includegraphics[width=0.7\columnwidth]{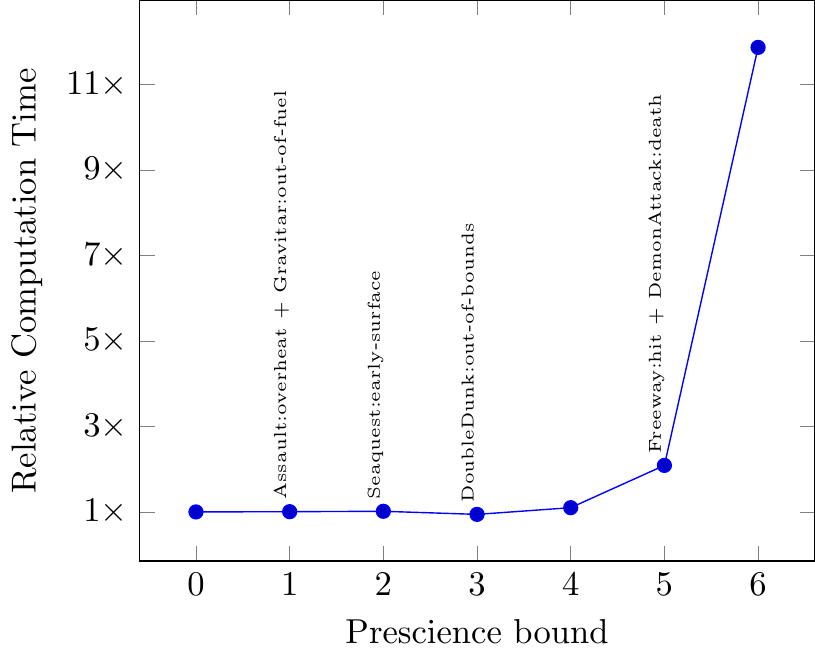}
  \caption{Compute cost of BPS and properties first satisfied, by prescience bound.}
  \label{fig:shield_compute}
\end{figure}

However, we emphasise that the relative computational cost of
\shield is exponential with respect to its prescience bound
Fig.~\ref{fig:shield_compute}.  This becomes a pressing issue when applying
\shield to games and properties that require a much larger prescience bound. 
An example of such a game and property is
$\texttt{\textbf{Bowling:no-hit}}$ where the agent needs to have a
prescience bound of hundreds of frames to avoid property violation
(Fig.~\ref{fig:bowling}).

\section{Conclusion}

This paper proposed \shield, the first explicit-state bounded prescience
shield for DRL agents in Atari games.  We have defined a library of 43
safety specifications that characterise ``safe behaviour''.  Despite the
fact that there is positive correlation between the reward and satisfaction
of these properties, we found that all of the top-performing DRL algorithms
violate these safety properties across all the games we have considered.  In
order to analyse these failures we have applied explicit-state model
checking to explore all possible traces induced by a trained agent.  An
analysis of these results suggests that most agents satisfy most of the
safety properties most of the time, but that (relatively) rare violations
remain.  We conjecture that this finding is due to the fact that the policy
of these agents is driven by an expected reward, which may be an ill-fit
when the goal is to obtain a worst-case guarantee.  Based on this
observation we propose a countermeasure that applies our explicit-state
exploration to implement bounded safety check we call \emph{bounded
prescience shield} to mitigate the unsafe behaviour of DRL agents.  We
demonstrate that our shield improves the overall safety of all agents across
all games at minimal computational cost, delivering the agents that are, to
the best of our knowledge, the safest agents available for ALE games.  We
observe that our safe agents obtain only marginally higher rewards on
average, which offers an explanation why DRL training does not prevent the
safety violations.

\begin{acks}
 This work is in part supported by UK NCSC, the HICLASS project
(113213), a partnership between the Aerospace Technology
Institute (ATI), Department for Business, Energy \& Industrial
Strategy (BEIS) and Innovate UK, and by the Future of Humanity Institute, Oxford.

\end{acks}

%%%%%%%%%%%%%%%%%%%%%%%%%%%%%%%%%%%%%%%%%%%%%%%%%%%%%%%%%%%%%%%%%%%%%%%%

%%% The next two lines define, first, the bibliography style to be 
%%% applied, and, second, the bibliography file to be used.
\balance

\if\doctype1
\bibliographystyle{ACM-Reference-Format} 
\bibliography{PAPER}
\else
\clearpage
\bibliographystyle{ACM-Reference-Format} 
\bibliography{PAPER}
\fi

%%%%%%%%%%%%%%%%%%%%%%%%%%%%%%%%%%%%%%%%%%%%%%%%%%%%%%%%%%%%%%%%%%%%%%%%
%% \clearpage
%% \input{appendix}
\end{document}